\documentclass[letterpaper, 10 pt, conference]{ieeeconf}  % Comment this line out if you need a4paper

\IEEEoverridecommandlockouts                              % This command is only needed if 
\usepackage{graphics} % for pdf, bitmapped graphics files
\usepackage{epsfig} % for postscript graphics files
\usepackage{mathptmx} % assumes new font selection scheme installed
\usepackage{times} % assumes new font selection scheme installed
\usepackage{amsmath} % assumes amsmath package installed
\usepackage{amssymb}  % assumes amsmath package installed
\usepackage{booktabs} 
\usepackage{cleveref} % for \cref references
\usepackage{multirow}
\usepackage{balance}
\usepackage{threeparttable}
\usepackage{fancyhdr}

\title{
\LARGE \bf
StateLinFormer: Stateful Training Enhancing Long-term Memory in Navigation
}

\author{Zhiyuan Chen$^{*}$, Yuxuan Zhong$^{*}$, Fan Wang$^{\dagger}$, Bo Yu, Pengtao Shao, Shaoshan Liu$^{\dagger}$, Ning Ding%
\thanks{$^{*}$Equal contribution.}%
\thanks{$^{\dagger}$Corresponding author. Email: {\tt\small  fanwang.px@gmail.com, shaoshanliu@cuhk.edu.cn}}%
\thanks{All authors are with the Shenzhen Institute of Artificial Intelligence and Robotics for Society, Shenzhen, China.
}
\thanks{\textit{Proceedings of the IEEE/RSJ International Conference on Intelligent Robots and Systems (IROS), 2026.
}}
}
\begin{document}

\maketitle
\lhead{Published as a conference paper at IROS 2026 }
\thispagestyle{empty}
\pagestyle{empty}

%%%%%%%%%%%%%%%%%%%%%%%%%%%%%%%%%%%%%%%%%%%%%%%%%%%%%%%%%%%%%%%%%%%%%%%%%%%%%%%%
\begin{abstract}

% Effective navigation intelligence necessitates integrating spatial recognition, common sense reasoning, and long-term memory to enable both immediate generalization and sustained adaptation. However, current approaches face a dilemma: modular systems tend to be inflexible, while large pre-trained models often lack the persistent memory crucial for thorough environmental adaptation. To enable long-horizon memory and online adaptation, we propose STREAM (\textbf{S}tateful \textbf{TR}aining of lin\textbf{E}ar \textbf{A}ttention with long-term \textbf{M}emory), a training protocol for linear attention architectures that maintains persistent memory across interaction batches instead of resetting them. Experiments in the ProcTHOR environment show that STREAM-trained models significantly outperform Transformer baselines with comparable parameter sizes and exhibit strong in-context adaptation over long horizons. These findings demonstrate that STREAM enables the model to adapt effectively to novel environments and achieve improved navigation performance.

Effective navigation intelligence relies on long-term memory to support both immediate generalization and sustained adaptation. However, existing approaches face a dilemma: modular systems rely on explicit mapping that must be rebuilt for each new environment, limiting scalability, while Transformer-based end-to-end models are constrained by fixed context windows, which restricts persistent memory across extended interactions. We introduce StateLinFormer, a linear-attention navigation model trained with a stateful memory mechanism that preserves recurrent memory states across consecutive training segments instead of reinitializing them at each batch boundary. This training paradigm effectively approximates learning on infinitely long sequences, enabling the model to achieve long-horizon memory retention. Experiments across both MAZE and ProcTHOR environments demonstrate that StateLinFormer significantly outperforms its stateless linear-attention counterpart and standard Transformer baselines with fixed context windows. Notably, as interaction length increases, persistent stateful training substantially improves context-dependent adaptation, suggesting an enhancement in the model's In-Context Learning (ICL) capabilities for navigation tasks.

\end{abstract}

%%%%%%%%%%%%%%%%%%%%%%%%%%%%%%%%%%%%%%%%%%%%%%%%%%%%%%%%%%%%%%%%%%%%%%%%%%%%%%%%
\section{INTRODUCTION}

Navigation is a fundamental capability for embodied agents and underpins a wide range of real-world applications, including autonomous inspection and surveillance \cite{remoteInspection, BeliefBehaviorGraph, Puthumanaillam2025EnhancingRN, CaveExploration}, service robotics \cite{ServiceRobotsHealthcare, rondoni2024navigationhospital}, and assistive home robots \cite{assistingolderadults, ServiceRobotsHealthcare, ElderlyCare}. Successful navigation requires integrating observations over extended time horizons to construct coherent spatial representations and make temporally consistent decisions. In realistic deployments, agents operate continuously within the same environment, where experience accumulates over time. Consequently, effective long-term memory is a central prerequisite for robust and adaptive navigation.

Classical navigation systems rely on explicit mapping and localization pipelines such as SLAM, which construct environment-specific maps for planning. While highly effective in structured settings, these approaches require rebuilding maps for each deployment and lack flexibility when scaling across diverse environments. More recent end-to-end approaches employ Transformer-based architectures, leveraging large-scale pretraining to encode rich semantic and spatial priors. Although such models demonstrate promising zero-shot and few-shot generalization, their fixed context windows constrain the amount of past information available at inference time. Consequently, this limitation leads to redundant exploration of previously visited areas while precluding the agent's ability to adapt to novel environmental layouts during sustained interaction.
% More importantly, these models are typically trained under an episodic paradigm in which memory is reset between trajectories. This training procedure implicitly assumes short, independent episodes and prevents the model from learning to exploit persistent cross-episode information, leading to fragmented global understanding and limited long-horizon adaptation.

To enable scalable long-horizon memory, we adopt a linear attention architecture that supports incremental state updates with constant computational complexity. In principle, such architectures can maintain persistent memory over arbitrarily long interaction streams. However, architectural capacity alone is insufficient. We argue that the core issue is not architectural capacity but optimization protocol. Many sequence modeling pipelines adopt a stateless training paradigm characterized by re-initializing the internal memory state at the start of each batch. Consequently, memory is primarily optimized under zero-initialized states and only fully exploited at inference time. This mismatch between architectural capability and training protocol constrains optimization to truncated segments and discourages the emergence of behaviors that depend on continuously evolving memory states. 

To bridge this gap, we introduce StateLinFormer, a linear-attention navigation model trained with a stateful memory mechanism that preserves hidden states across consecutive training batches instead of resetting them. By maintaining memory continuity during optimization, historical context conditions gradient updates beyond a single truncated segment. As a result, model parameters are optimized under a distribution of accumulated memory states rather than independently re-initialized segments. This training paradigm effectively approximates learning over an infinite interaction stream instead of disjoint truncated sequences.
Crucially, preserving persistent memory alters the statistical structure of the training signal. This exposure also increases the burstiness of training signals, which facilitates the emergence of in-context learning (ICL) behavior \cite{chan2022data} and promotes stronger online adaptation in novel environments.

Experiments in both MAZE and ProcTHOR environments demonstrate that StateLinFormer consistently outperforms its stateless linear-attention counterpart under identical architectures and parameter counts. In addition, it surpasses parameter-matched Transformer baselines with fixed context windows. Notably, these empirical results confirm that persistent stateful training leads to substantial gains in long-horizon performance and in-context adaptation, highlighting the importance of aligning training with continuous interaction.

Our contributions are summarized as follows:
\begin{itemize}
    \item \textbf{Stateful training of linear attention model for navigation.} We propose a stateful training paradigm for linear-attention navigation models that preserves memory states across consecutive training batches instead of resetting them. This approach approximates optimization over effectively unbounded interaction sequences, aligning training with the continual nature of embodied deployment and enabling persistent long-horizon memory.
    \item \textbf{Emergent In-Context Learning via stateful training.} 
    We show that preserving memory continuity during training leads to sustained performance improvements as interaction context increases, without parameter updates. This behavior indicates enhanced in-context learning and cross-episode adaptation induced by stateful training.
    \item \textbf{Continual Object Navigation (CON) benchmark.} We introduce the Continual Object Navigation (CON) benchmark to evaluate adaptation under persistent interaction within the same environment. CON requires agents to accumulate and reuse experience over extended interaction horizons rather than operating on independently truncated segments. To support this protocol, we construct continuous interaction streams that preserve temporal continuity within each environment.

\end{itemize}
\section{RELATED WORK}

\subsection{Explicit Mapping-Based Navigation}
% Explicit mapping-based navigation, anchored by Simultaneous Localization and Mapping (SLAM), remains a dominant paradigm in robotics. Historically, SLAM frameworks have been categorized by their sensor modalities: Visual SLAM (e.g., \cite{ORB-SLAM3, engel2016directsparseodometry, jiang2025monoslamrobustmonocularslam, 10238802, Yan_2024_CVPR}) and LiDAR SLAM (e.g., \cite{zhang2014loam, shan2020lio, pan2024pin, cui2022bow3d}). LiDAR-based systems are valued for their high spatial resolution and robustness in indoor environments, while Visual SLAM offers a lower-cost alternative that captures rich texture but is sensitive to lighting variations and "feature-starvation" in monochromatic spaces. However, these classical models both lack the contextual awareness necessary for navigating complex, dynamic environments\cite{rs14236033}.
Classical navigation paradigms often rely on explicit spatial representations, most prominently through Simultaneous Localization and Mapping (SLAM). Among these, LiDAR-based systems \cite{zhang2014loam, shan2020lio, pan2024pin, cui2022bow3d} provide high-precision geometric reconstruction and robustness under diverse lighting conditions, while Visual SLAM (vSLAM) \cite{ORB-SLAM3, engel2016directsparseodometry, jiang2025monoslamrobustmonocularslam, 10238802, Yan_2024_CVPR} offers a cost-effective alternative by leveraging dense visual and photometric cues.
Despite their fidelity in metric reconstruction, explicit mapping approaches face three fundamental challenges in intelligent navigation. First, although they produce accurate geometric maps, they lack intrinsic semantic and task-level representations. Second, their modular pipelines are susceptible to cascading errors. Third, their reliance on explicit mapping and predefined components limits flexibility.

% Classical navigation paradigms rely heavily on explicit spatial representations, primarily through Simultaneous Localization and Mapping (SLAM). These frameworks are traditionally bifurcated by sensor modalities: LiDAR-based systems \cite{zhang2014loam, shan2020lio, pan2024pin, cui2022bow3d}, which provide high-precision geometric structures and robustness in diverse lighting, and Visual SLAM (vSLAM) \cite{ORB-SLAM3, engel2016directsparseodometry, jiang2025monoslamrobustmonocularslam, 10238802, Yan_2024_CVPR}, which offers a cost-effective alternative by leveraging rich textual and photometric information.
% Despite their high fidelity in metric reconstruction, explicit mapping approaches face two major hurdles in intelligent navigation. First, they often lack semantic and contextual awareness; while they can reconstruct a "cloud of points", they typically rely on geometric reconstruction and require additional modules to incorporate semantic or task-level reasoning. Second, these modular systems suffer from cascading errors. 

\subsection{Transformer Models for Navigation}

Recent advances in embodied navigation increasingly adopt Transformer-based architectures due to their strong sequence modeling capacity and ability to integrate multimodal observations, actions, and language instructions enabled by large-scale pretraining. The research landscape is typically categorized by task scenarios. Object-goal navigation \cite{shah2023vint, bar2024navigation, NEURIPS2024_c7286145, sridhar2023nomad} focuses on locating a target object category in previously unseen environments, requiring exploration and semantic reasoning. Social navigation \cite{10610371, 10802716, 11128004} emphasizes human-centric environments and multi-agent interaction modeling. Vision-Language Navigation (VLN) \cite{Anderson_2018_CVPR, liang2023contextaware, Wang_2019_CVPR, moudgil2021soat, ehsani2024spoc, Kamath_2023_CVPR, Liu_2024_CVPR} requires cross-modal alignment between natural language instructions and visual observations. Beyond task formulation, prior work can also be organized by learning paradigm. Some approaches adopt supervised learning with labeled action sequences \cite{shah2023vint, sridhar2023nomad, Liu_2024_CVPR, NEURIPS2024_c7286145}, others rely on imitation learning from expert demonstrations \cite{moudgil2021soat, Wang_2019_CVPR, Kamath_2023_CVPR, ehsani2024spoc}, while reinforcement learning methods optimize policies through environment interaction and reward signals \cite{liang2023contextaware, pmlr-v270-zeng25a, Zhu_2025_ICCV, elawady2025relic, gupta2025memo}.

Despite this diversity in task settings and learning paradigms, most Transformer-based navigation models are trained with fixed context windows and reset hidden states between trajectories, limiting their ability to accumulate persistent memory over extended interaction streams.

\subsection{Memory-Based Adaptation}

Memory is a cornerstone for efficient navigation, as it allows agents to leverage past experiences to optimize current decision-making. 
In recent years, several works have introduced explicit memory mechanisms to extend effective context horizons in Transformer-based navigation models. 
ReLIC \cite{elawady2025relic} extends the effective context window to tens of thousands of steps via learnable key-value memory vectors, while Memo \cite{gupta2025memo} compresses long histories into summary tokens for efficient retrieval.
While these memory-based methods demonstrably enhance adaptive capabilities, their memory mechanisms are typically trained under settings where hidden states are reset between training trajectories. As a result, memory does not persist across training batches, limiting exposure to continuously evolving interaction streams.
This reflects a broader challenge in sequence modeling: maintaining information across sequence boundaries. Early work, such as Katrompas et al. \cite{katrompas2021enhancing}  explored stateful training for LSTM-based sequence classification tasks, demonstrating that preventing state resets can improve long-term dependency modeling. 
However, such stateful optimization has not been systematically investigated in long-horizon embodied navigation, particularly with scalable linear-attention architectures.

\section{Methodologies}

\subsubsection{Problem Formulation}
We formulate navigation as a partially observable Markov decision process (POMDP), defined by the tuple
% \[
% (\mathrm{S}, \mathrm{A}, \mathrm{P}, \mathrm{R}, \mathrm{O}, \mathrm{Z}).
% \]
\[
(\mathrm{S}, \mathrm{A}, \mathrm{P}, \mathrm{O}, \mathrm{Z}).
\]
Here, $s_t \in \mathrm{S}$ denotes the underlying environment state at time $t$, which includes the agent's pose and the unobserved global scene layout; $a_t \in \mathrm{A}$ denotes the navigation action; $\mathrm{P}(s_{t+1} \mid s_t, a_t)$ defines the environment dynamics.
% and $\mathrm{R}(s_t, a_t)$ specifies the task-dependent reward. 
The agent receives a partial observation $o_t \in \mathrm{O}$ generated according to the observation model $\mathrm{Z}(o_t \mid s_t)$, reflecting limited sensor coverage and occlusions. We choose to omit explicit consideration of rewards, since rewards can typically be derived from the state or observation. 
A navigation trajectory $\tau$ of length $T$ is defined as a sequence of prompts, observations, and actions:
\[
\tau = \{(p_t, o_t, a_t)\}_{t=1}^T
\]
, where $p_t$ denotes the conditioning prompt (e.g., task instruction or goal direction). The trajectory terminates upon reaching the navigation goal or exceeding a maximum time horizon. 
Due to partial observability and the goal-conditioned nature of the task, we consider policies that rely on the prompt and interaction history:
\begin{align}
    &\hat{a}_t \sim \pi_{\theta}(\cdot \mid q_t), \nonumber\\
    &q_t = (p_{t-\Delta t}, o_{t-\Delta t}, a_{t-\Delta t}, \dots, p_{t-1}, o_{t-1}, a_{t-1}, p_t, o_t) \nonumber
\end{align}
, where $q_t$ represents the combined context of the task prompt and the temporal observation-action sequence, and $\theta$ denotes the model parameters. 

Our objective is to learn a navigation policy that approximates the oracle policy by minimizing the imitation learning loss over a dataset of expert trajectories $\mathrm{E}$. Formally, we optimize $\theta$ by minimizing the negative log-likelihood (NLL) of the ground-truth expert actions:
\begin{align}
\mathrm{L}(\theta) = - \mathbb{E}_{\tau \sim \mathrm{E}} \left[ \sum_{t=1}^{T} \log \pi_\theta(a_t \mid q_t) \right].
\end{align}
By incorporating $p_t$ into the sequence $q_t$, the policy $ \pi_\theta$ learns to bridge the gap between high-level instructions and low-level embodied control, effectively mapping the latent spatial structure and task requirements to optimal navigation actions.

\subsection{Model Architecture}
% The architecture of our model, STREAM, follows a modular design inspired by the SPOC framework \cite{}, consisting of a goal encoder, a goal-conditioned visual encoder, and an action decoder. At each time step $t$, the agent receives a text instruction $\mathrm{G}$ and visual observations from two egocentric cameras $\mathrm{F}_{nav}^t, \mathrm{F}_{manip}^t$. 
% The goal encoder $\mathrm{E}_{goal}$ maps $\mathrm{G}$ to contextualized tokens $\mathbf{g}$. 
% The visual encoder $\mathrm{E}_{visual}$ integrates these tokens with image patch embeddings through a Transformer-based fusion module, outputting an observation representation $\mathbf{o}_t \in \mathbb{R}^d$. 
% While we adopt the perception backbone from SPOC to ensure robust multi-modal grounding, we replace the standard Transformer decoder with a Linear Attention-based memory backbone and introduce. Unlike SPOC, which uses cross-attention to condition the decoder on a fixed goal, we integrate the goal information directly into the input stream at the encoder level. This design choice supports long-horizon sequences that may contain multiple sequential tasks.

\begin{figure}[htbp]
    \includegraphics[width=0.95\linewidth]{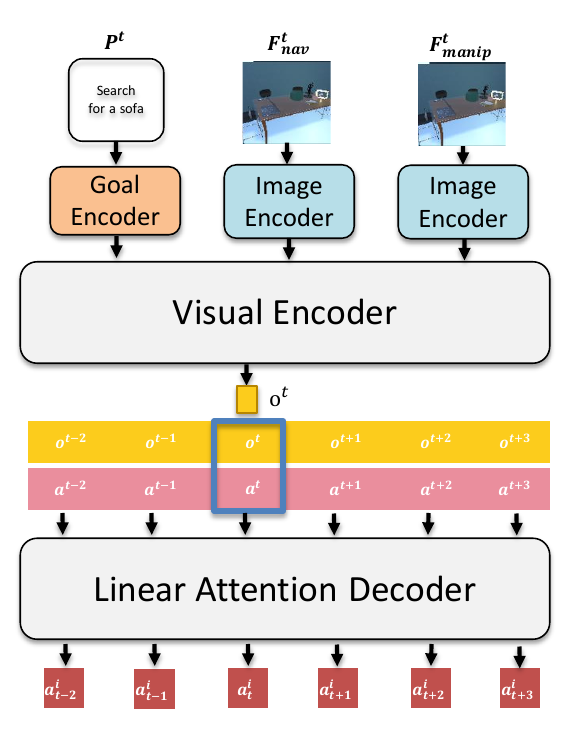}   
    \caption{The model structure of StateLinformer}
    \label{fig:structure}
    \vspace{-0.15in}
\end{figure}

StateLinFormer adopts a modular encoder-decoder structure for navigation, inspired by the SPOC framework \cite{ehsani2024spoc} and illustrated in Fig.~\ref{fig:structure}. At each time step $t$, the agent receives a text instruction $P_t$ and visual observations from navigation and manipulation cameras, $F_{nav}^t$ and $F_{manip}^t$. The architectural processing pipeline is defined as follows:

First, $P_t$, $F_{\text{nav}}^t$, and $F_{\text{manip}}^t$ are processed by the goal encoder $\mathrm{E}_{\text{goal}}$ and image encoder $\mathrm{E}_{\text{image}}$ to obtain representations $p_t$, $f_{\text{nav}}^t$, and $f_{\text{manip}}^t$.
The goal-conditioned visual encoder $\mathrm{E}_{visual}$ then integrates these features to produce a unified observation representation:
\begin{align}
    & p_t = \mathrm{E}_{\text{goal}}(P_t),\\
    & f_{nav}^t = \mathrm{E}_{\text{image}}(F_{nav}^t),\\
    & f_{manip}^t = \mathrm{E}_{\text{image}}(F_{manip}^t), \\
    & o_t = \mathrm{E}_{visual}(f_{nav}^t, f_{manip}^t, p_t).
\end{align}

For the action decoder $\mathrm{D}$, we replace the standard Transformer backbone with a linear-attention architecture. To support trajectories containing multiple instructions, we remove explicit goal cross-attention within the decoder and instead condition actions directly through the integrated observation representation $o_t$.

At each step, the decoder takes the current observation $o_t$ and previous action $a_{t-1}$, while maintaining an internal memory state $M_{t-1}$. The input is defined as:
\begin{align}
    x_t &= o_t \oplus a_{t-1}, \\
    h_t, M_t &= \mathrm{D}(x_t, M_{t-1}),
\end{align}
where $M_t \in \mathbb{R}^{d \times d}$ denotes the persistent memory state.

Following the standard kernelized linear attention formulation, the memory is updated as
\begin{align}
    M_t &= M_{t-1} + \phi(k_t) v_t^\top, \\
    h_t &= \phi(q_t)^\top M_t,
\end{align}
where $\phi(\cdot)$ is a feature mapping function. 
This formulation maintains a fixed-size memory state and constant computational cost per time step with respect to sequence length.

Finally, the action distribution is predicted as
\begin{equation}
    \pi_t = \text{Softmax}(\text{MLP}(h_t)).
\end{equation}
Importantly, while the architectural form of the memory update remains standard, the key contribution of StateLinFormer lies in how the memory state $M_t$ is handled during training, as described in the following section.

\subsection{Stateful Training of linear Attention with long-term Memory}
Linear attention models maintain an explicit memory state that incrementally aggregates past information. 
% Given input embeddings $x_{t}$
% , the memory state is updated recurrently as:
% \begin{align}
% M_t = M_{t-1} + \phi(k_t) v_t^\top
% \end{align}
% where $k_{t}$ and $v_{t}$ are linear projections of $x_{t}$, and 
% $\phi(\cdot)$ is a feature map ensuring linear complexity.
Existing training pipelines typically adopt stateless training. During training, the memory state is re-initialized at the start of each batch:
$M^{(b)}_0 = 0$.

Instead of resetting memory, StateLinFormer maintains the continuity of the memory state across successive training batches. Specifically, for a continuous trajectory divided into batches of length $T$, StateLinFormer initializes the memory of batch $b+1$ using the final state of batch $b$:
\begin{align}
    M^{(b)}_T \rightarrow M^{(b+1)}_0
\end{align}
where $M^{(b)}_T$ is the terminal state of the $b$-th batch and $M^{(b+1)}_0$ is the initial state for the subsequent batch. The training loss is computed per batch as usual:
\begin{align}
\mathrm{L} = \sum_{t=1}^{T} \ell(\pi(a_t \mid h_t), a_t)
\end{align}

\begin{figure*}[t]
\centering
\includegraphics[width=0.95\linewidth]{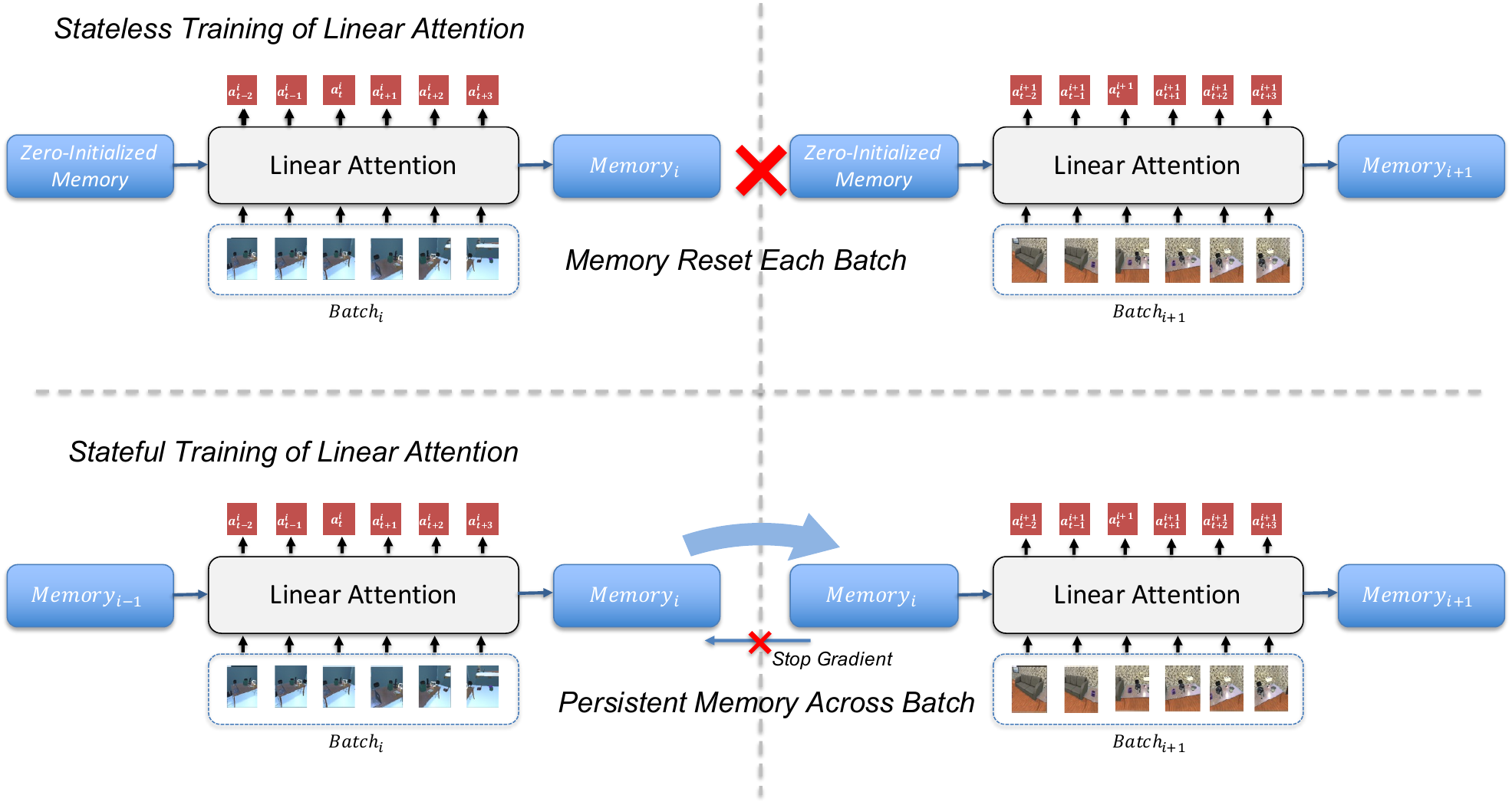}
\caption{The framework of StateLinFormer. Top: Conventional stateless training with memory reset each batch. Bottom: StateLinFormer, which maintains persistent memory across batches to support long-horizon adaptation.
}
\label{fig:framework}
\end{figure*}

By propagating the memory state forward, StateLinFormer ensures the model parameters are trained under consecutive memory that reflects long-range temporal dependencies and past exploration. By propagating the memory state forward across batches, StateLinFormer approximates forward-state exposure for infinitely long sequences, while truncating gradients within each batch. The framework of StateLinFormer is shown in Figure~ \ref{fig:framework}.
% As a result, the model is repeatedly optimized under memory states that encode arbitrarily long histories, effectively exposing the parameters to long-range temporal dependencies without explicitly unrolling long sequences.

\subsection{Optimization Perspective}
While stateful training can be described operationally as propagating memory states across batches, its primary impact lies in how it reshapes the optimization objective.

Under conventional stateless training, the memory state is re-initialized at the beginning of each batch, typically as $M_0 = 0$. Consequently, model parameters are optimized under a degenerate distribution over memory states concentrated at zero. Formally, the objective can be expressed as:
\begin{equation}
    \min_{\theta} \mathbb{E}_{\tau \sim \mathcal{D}} L(\theta; M_0 = 0),
\end{equation}

Stateful training preserves memory continuity across consecutive batches. Let the recurrent memory update be defined as
\begin{equation}
    M_t = f_{\theta}(M_{t-1}, x_t),
\end{equation}
where $x_t$ denotes the input at time step $t$. By propagating $M_T^{(b)} \to M_0^{(b+1)}$, the model is optimized under memory states generated by its own long-horizon dynamics. The training objective can then be approximated as:
\begin{equation}
    \min_{\theta} \mathbb{E}_{\tau \sim \mathcal{D}, M \sim d_{\theta}} L(\theta; M),
\end{equation}
where $d_{\theta}$ denotes the distribution over memory states induced by the model's recurrent dynamics along extended trajectories.
Under mild ergodicity assumptions, such recurrent dynamics induce a stationary distribution over memory states. Stateful training exposes the model parameters to samples drawn from this distribution, while truncating gradients within each batch for computational tractability. 
This shift aligns the training objective with the continual nature of deployment. Instead of optimizing for short, independently reset segments, the model is optimized under the evolving memory states it will encounter during sustained interaction. 

\section{Experiments}

\subsection{Experimental setup}

\textbf{Continual object navigation task (CON).}  Building upon prior formulations of object navigation~\cite{sun2024survey} and multi-object navigation (MultiON)~\cite{wani2020multion}, we introduce Continual Object Navigation (CON), a task setting that extends the agents to operate under sustained conditions. Compared to MultiON, CON differs in three aspects: (1) the generation of an unbounded sequence of goals (permitting goal repetition across episodes), (2) the absence of advance goal specification (the agent receives subsequent objectives only upon completing the prior task), and (3) sustained environmental continuity (the agent remains situated within the same environment throughout the entire sequence). Conceptually, CON may be interpreted as stitching individual navigation episodes into a prolonged sequence, where episodic boundaries are retained but environmental context persists uninterrupted. CON simulates a highly practical scenario that aligns with the principles of continual or lifelong learning: a household robot is deployed in an unfamiliar room, and gradually becomes familiar with and adapts to the room as it completes various tasks.

\textbf{Dataset.}
To train our models at scale, we utilize procedurally generated environments from two distinct domains, both representing indoor settings: a grid-based Maze and the visually realistic ProcTHOR \cite{10.5555/3600270.3600703}. Our Maze environment is designed to capture pixel-style spatial semantic features. Closely related to the settings described in \cite{pavsukonisevaluating2023} and \cite{wang2024architect}, it is generated on a  $15 \times 15$ grid world. As demonstrated by previous studies \cite{pavsukonisevaluating2023, wang2024benchmarking, pleines2025memory}, procedurally generated mazes pose substantial challenges for transition prediction due to their inherent stochasticity and the agent’s limited sensory input (i.e., partial observability).
In contrast, ProcTHOR serves as a more realistic 3D indoor counterpart. Built within the AI2-THOR \cite{Kolve2017AI2THORAI} simulator and populated with annotated 3D assets from Objaverse \cite{Deitke_2023_CVPR}, it generates egocentric visual inputs at a 
$394 \times 224$ resolution.
% Across these environments, we formulate two training variants based on trajectory types: (i) single-episode trajectories, where each trajectory contains a single navigation goal with variable length, and (ii) cross-episode trajectories, where each trajectory may include multiple sequential goals within a fixed length to support continual navigation scenarios (as formalized by the Continual Object Navigation task below). 
% In Maze, we focus on cross-episode trajectories to emphasize long-horizon memory stress testing. In ProcTHOR, we generate both formats to evaluate generalization under both episodic and continual settings.
To facilitate the Continual Object Navigation (CON) task, we develop a data generation protocol that emphasizes sustained interaction. Each training sequence is composed of multiple navigation goals delivered in succession within a persistent environment. Detailed statistics of the generated datasets are summarized in Table~\ref{tab:data_statistics}.

% \begin{table}[htbp] % 
%     \centering % 
%     \caption{A summary of data distribution across the training datasets.}
%     \label{tab:data_statistics} % 
%     \resizebox{\linewidth}{!}{% % 
%     \begin{tabular}{l|ccccc}
%     \toprule
%     Dataset & \# envs ($|\mathrm{E}|$) & Len. Traj. & \# Traj. & \# time steps \\
%     \midrule
%     \textit{MAZE} \\
%     \quad Cross-Episode & 1k & 2000 & 50K & 10000M \\
%     \midrule
%     \textit{ProcTHOR} \\
%     \quad Single-Episode & 1k & $<$ 500 & 10K & 1000M  \\
%     \quad Cross-Episode &  & 2000 & 1K & 1000M \\
%     \bottomrule
%     \end{tabular}%
%     }
% \end{table}

\begin{table}[htbp] % 
    \centering % 
    \caption{A summary of data distribution across the training datasets.}
    \label{tab:data_statistics} % 
    \resizebox{\linewidth}{!}{% % 
    \begin{tabular}{l|ccccc}
    \toprule
    Dataset & \# envs ($|\mathrm{E}|$) & Len. Traj. & \# Traj. & \# time steps \\
    \midrule
    MAZE & 1k & 2000 & 50K & 100M \\
    \midrule
    ProcTHOR & 1k & 2000 & 5K & 10M \\
    \bottomrule
    \end{tabular}%
    }
\end{table}

% \textbf{Model Structure Details.} 
% We utilize RWKV-7 \cite{peng2025rwkv} as our backbone linear attention architecture to model state transitions. Specifically, the model is configured with 18 layers and a hidden dimension of 512, totaling approximately 0.2 billion parameters, with 32 attention heads per layer dedicated to processing spatial representations.
% For feature extraction, both the goal encoder $\mathrm{E}_{\text{goal}}$ and the image encoder $\mathrm{E}_{\text{image}}$ utilize SigLIP pre-trained weights, while the visual encoder $\mathrm{E}_{\text{visual}}$ follows the architecture design of SPOC. 

% \textbf{Training Details.} During training, the parameters of $\mathrm{E}_{\text{goal}}$ and $\mathrm{E}_{\text{image}}$ are kept frozen. The training process is initialized with a learning rate of $\mathrm{1e}^{-4}$, which follows a linear decay schedule throughout the training duration. We employ a batch size of 4 for all experiments. The model is trained on 8 NVIDIA A800 GPUs. We train our model for a total of 10 epochs.

% \paragraph{}

\textbf{Baselines.} 
To evaluate the effectiveness of our approach, we compare it against the following baselines:
\begin{itemize}
    \item \textbf{Stateless Training Linear Attention }: An ablation of StateLinFormer with an identical architecture and parameter count, differing only in the training protocol. The memory states are reset at the beginning of each training batch, resulting in standard stateless optimization.
    % \item \textbf{Transformer} \cite{NIPS2017_3f5ee243}: A controlled architectural baseline obtained by replacing the linear-attention backbone in StateLinFormer with a standard Transformer decoder employing global self-attention. Its maximum context length is set to 100.
    \item \textbf{SPOC} \cite{ehsani2024spoc}: SPOC follows an architectural pipeline highly similar to ours, employing a Transformer decoder in place of linear attention. From an architectural perspective, SPOC can be viewed as a Transformer variant of our model. To ensure a comprehensive evaluation, we assess SPOC under two distinct settings: (i) \textbf{SPOC-10M}, which is re-trained from scratch on our 10M-frame dataset, serving as a direct Transformer-based architectural baseline; and (ii) \textbf{SPOC-Pretrained}, the original pre-trained model to provide a large-scale performance reference. In both settings, we maintain its default configuration with a context window of 100 steps.
    % \item \textbf{PoliFormer} \cite{pmlr-v270-zeng25a}: PoliFormer utilizes an architectural pipeline similar to SPOC but is optimized through a large-scale reinforcement learning (RL) framework. Due to its heavy reliance on massive computational resources and the specific data requirements of RL-based optimization, we evaluate PoliFormer exclusively using its original released checkpoint with a context length of 128, serving as a high-performance reference for large-scale pre-training (\textbf{PoliFormer-Pretrained}).
    % \item \textbf{Dreamer-V3} \cite{hafner2025mastering}: A model-based reinforcement learning approach that learns a latent world model and performs planning via imagination in latent space. Dreamer maintains recurrent latent states to capture temporal dependencies and has demonstrated strong performance in partially observable environments. We use the standard configuration adapted to our navigation settings.
    % \item \textbf{VINT} \cite{shah2023vint}: A Transformer-based foundation model for visual navigation pre-trained on diverse, cross-embodiment robotic datasets. ViNT learns general navigational affordances via an image-goal reaching objective, enabling zero-shot generalization across novel robots and efficient adaptation to new goal modalities. We evaluate the model using a context length of 5, as in its original design.
\end{itemize}
Due to differences in environmental characteristics and the computational requirements of various algorithms, we select specific subsets of baselines for each benchmark. 
On the Maze benchmark, we compare StateLinFormer against Stateless training linear attention and SPOC-10M to assess performance. We exclude the pre-trained versions of SPOC-Pretrained here due to the significant domain discrepancy between its visually realistic training environments and our grid-based Maze; direct zero-shot evaluation on such distinct modalities would be inherently unfair and uninformative.
On ProcTHOR, we include SPOC-Pretrained as a large-scale reference model. Since it is trained on substantially larger datasets, we directly evaluate its released weights without retraining, while all other baselines are retrained under our training dataset.

Furthermore, SPOC is designed with fixed context windows and is trained exclusively on single-episode trajectories, without exposure to long-horizon cross-episode interaction streams. To ensure a fair evaluation under the CON setting, we reset their context memory after each task completion. Maintaining an unbounded interaction stream without resetting would exceed their architectural context limits and lead to severe performance degradation unrelated to their intended design. In contrast, StateLinFormer is evaluated under continuous interaction streams consistent with their training configurations.
To ensure a fair comparison, all baseline models are matched in capacity with approximately 0.2 billion parameters.

\textbf{Evaluation Details.} All evaluations are conducted in unseen environments. By default, we use an evaluation set of size $|\mathrm{E}|=16$ for performance assessment, conducted across 16 distinct test environments. In each environment, the agent is allowed a maximum of 5000 execution time steps, within which multiple task instructions are provided sequentially. Each instruction is capped at a maximum of 500 time steps in Maze and 1000 time steps in ProTHOR. We employ Success Rate and Steps to Goal as our primary evaluation metrics. Success Rate represents the proportion of tasks where the agent successfully reaches the target specified by the instruction. Steps to Goal measures the number of execution steps taken to complete a task; for failed attempts, this value is set to the maximum allowable limit (500 in Maze and 1,000 in ProcTHOR).

% We exclude the pre-trained versions of SPOC-Pretrained and PoliFormer-Pretrained here due to the significant domain discrepancy between their visually realistic training environments and our grid-based Maze; direct zero-shot evaluation on such distinct modalities would be inherently unfair and uninformative.
% On ProcTHOR, we evaluate StateLinFormer against Stateless training linear attention, SPOC-10M, SPOC-Pretrained, and PoliFormer-Pretrained. 

% It is important to note that PoliFormer is included only as a pre-trained baseline. We do not re-train PoliFormer on our 10M-frame dataset because its large-scale reinforcement learning paradigm relies on massive computational resources and data volumes that far exceed our controlled experimental setting; furthermore, its performance and stability cannot be guaranteed under such a restricted data scale.

\subsection{Maze}

%\begin{table}[htbp]
%    \centering 
%    \caption{Comparison of the Performances in Mazes on CON Tasks. }
%    \label{tab:maze_res} 
%    \begin{tabular}{ccc}
%    \toprule
%    Model & Success Rate  & Steps to Goal  \\
%    \midrule
%    StateLinFormer (Stateless)   & 0.64 & 249  \\
%    StateLinFormer (Stateful) & 0.77 &  189 \\
%    SPOC-10M                   & 0.68 & 239  \\
%    \bottomrule
%    \begin{tablenotes}
%    Total frames of training in Maze are 10 M.
%    \end{tablenotes}
%    \end{tabular}
%\end{table}

\begin{table}[htbp]
    \centering
    \begin{threeparttable}
        \caption{Performance Comparison in Maze (CON Tasks).}
        \label{tab:maze_res}
        \small % Slightly smaller for a cleaner look
        \begin{tabular}{lcc} % Changed to lcc for better alignment
            \toprule
            Model & Success Rate $\uparrow$ & Steps to Goal $\downarrow$ \\
            \midrule
            StateLinFormer (Stateless) & 0.64 & 249 \\
            StateLinFormer (Stateful)  & \textbf{0.77} & \textbf{189} \\
            SPOC-10M                   & 0.68 & 239 \\
            \bottomrule
        \end{tabular}
        \begin{tablenotes}
            \item \textit{Note:} All models were trained for 10M frames in the Maze environment.
        \end{tablenotes}
    \end{threeparttable}
\end{table}

As shown in Table~\ref{tab:maze_res}, StateLinFormer consistently outperforms its stateless counterpart, and the Transformer-based baseline suggests that persistent memory exposure during training plays a critical role. These results provide preliminary evidence that state persistence during training improves performance in relatively simple, grid-based navigation settings.
% Since the two models share identical architectures and differ only in whether memory states are preserved across training batches, this comparison isolates the impact of stateful optimization. The performance gap indicates that training under persistent memory states correlates with improved generalization and long-horizon reasoning. 

\subsection{ProcTHOR}

Given the limited spatial and semantic complexity of the Maze environment, we further assess the effectiveness of our approach in the more realistic and semantically rich ProcTHOR environment. The results are reported in Table~\ref{tab:procthor_res}.

\textbf{Stateful training enhances linear attention under long-horizon interaction.}
Our results demonstrate that StateLinFormer consistently outperforms all baseline models in the ProcTHOR environment. Notably, it surpasses SPOC-Pretrained despite the latter being trained on significantly larger data volumes (40M frames). These findings further validate that StateLinFormer remains highly effective in semantically complex environments, proving that persistent memory is key to handling long-horizon tasks with high semantic demand.

\begin{table}
\caption{Performance comparison on CON tasks in ProcTHOR.}
\label{tab:procthor_res} 
\begin{tabular}{cccc}
\hline
Model                                                                & Success Rate$\uparrow$ & Steps to Goal$\downarrow$ & Total Frames \\ \hline
\begin{tabular}[c]{@{}c@{}}StateLinFormer\\ (Stateless)\end{tabular} & 0.420        & 669           & 10M          \\
\begin{tabular}[c]{@{}c@{}}StateLinFormer \\ (Stateful)\end{tabular} & \textbf{0.580}        & \textbf{496}           & 10M          \\
SPOC-10M                                                             & 0.479        & 630           & 10M          \\
SPOC-Pretrained                                                      & 0.566        & 525           & 40M          \\ \hline
\end{tabular}
\end{table}

\begin{figure}[htbp]
    \includegraphics[width=0.95\linewidth]{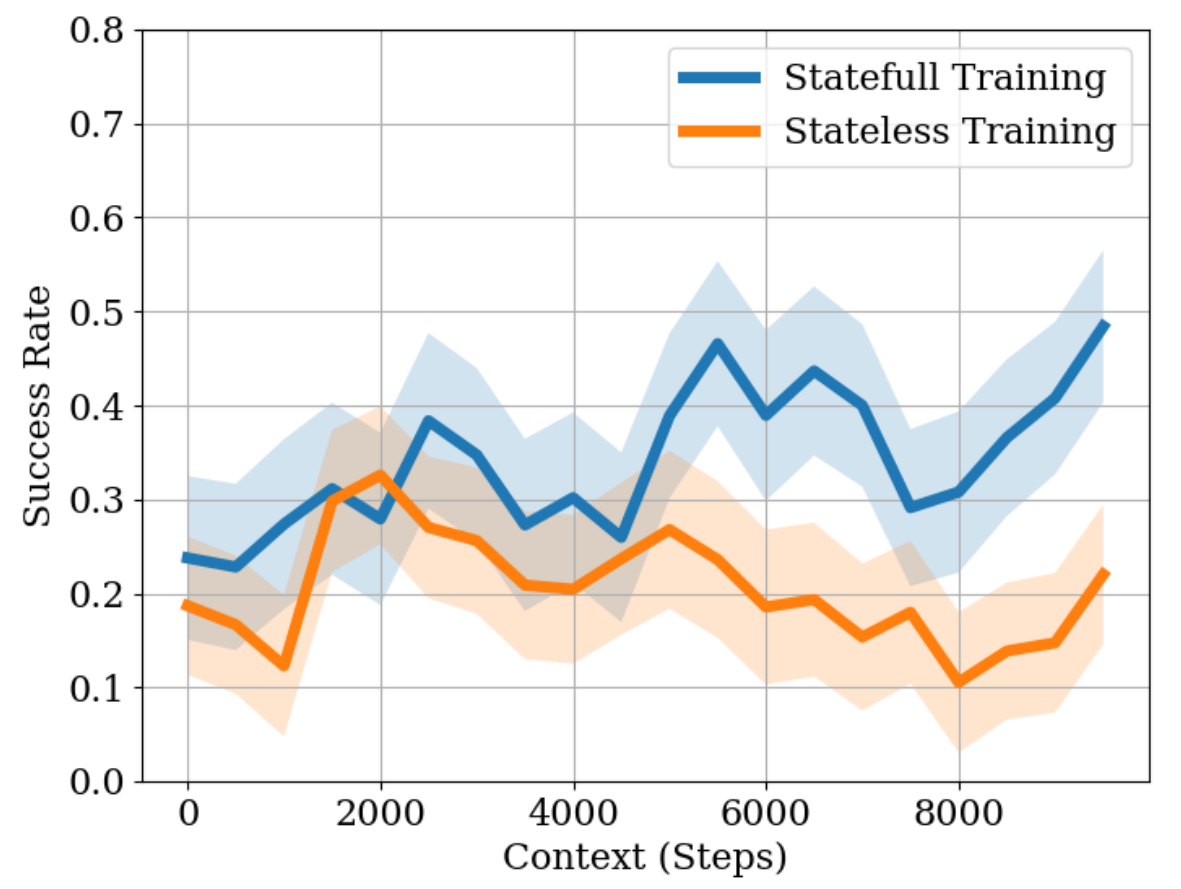}   
    \caption{Comparison of Success Rates between Stateful and Stateless Training across Increasing Context Lengths}
    \label{fig:stream_icl}
    \vspace{-0.15in}
\end{figure}

\textbf{Emergent In-Context Learning from Optimization Under Persistent Memory States.} To systematically evaluate the model's capacity for online adaptation, we measure the success rate as a function of context length (interaction steps) across unseen test environments, as illustrated in Fig.~\ref{fig:stream_icl}. The results reveal a striking difference between Stateful training and Stateless training in linear attention as the context length increases. While the stateless model exhibits marginal gains at short contexts and even degrades at longer horizons, the stateful model demonstrates a consistent improvement in success rate as the context grows. This trend strongly suggests that stateful training enables the model to accumulate and utilize information across extended temporal contexts, leading to improved performance without parameter updates. This behavior is characteristic of in-context learning: performance improves solely through conditioning on accumulated experience within the current interaction stream. Importantly, this gain is not due to architectural changes but arises purely from the stateful training protocol, which preserves memory continuity across batches. 

\subsection{Empirical Evidence for Stable Memory Regimes.}

To empirically examine the induced memory distribution $d_{\theta}$, we analyze the temporal statistics of the memory state norm under both training protocols. Specifically, we compute the relative standard deviation (RSD) of the memory norm over long trajectories.

\begin{figure}[htbp]
    \includegraphics[width=0.95\linewidth]{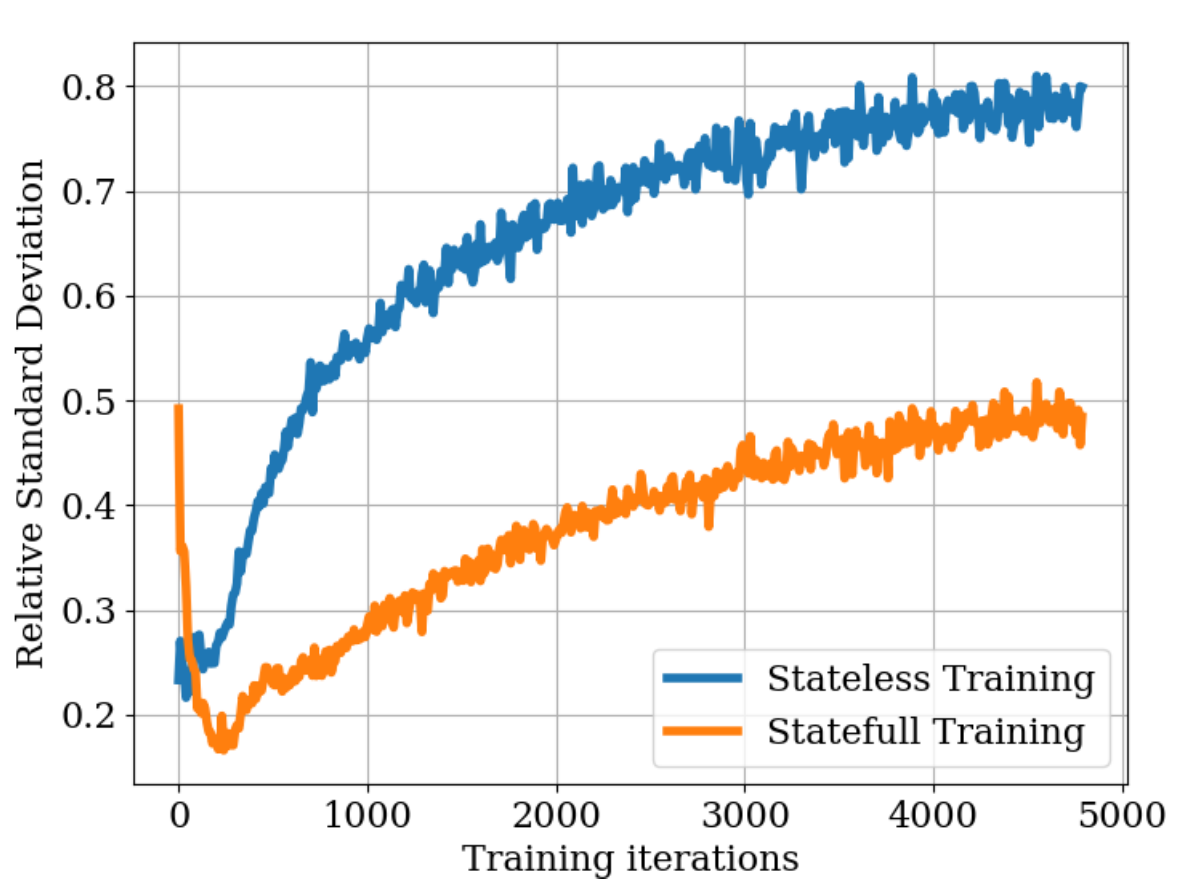}   
    \caption{Relative Standard Deviation (RSD) of Memory Norms: Stateful vs. Stateless Training.}
    \label{fig:memory_rsd}
    \vspace{-0.15in}
\end{figure}

As shown in Figure~\ref{fig:memory_rsd}, stateful training consistently exhibits lower RSD compared to stateless training. 
These findings are consistent with the optimization perspective described above. Stateless training optimizes parameters under a degenerate distribution concentrated near $M=0$, leading to transient memory dynamics. Stateful training, by contrast, exposes the model to memory states generated by its own long-horizon evolution, yielding empirical behavior consistent with sampling from an approximately stationary distribution $d_{\theta}$.

% \subsection{Discussion}
%/section{LIMITATIONS}
\section{Limitations}

While StateLinFormer demonstrates consistent advantages under persistent memory training, several limitations remain. First, although we provide an optimization-based interpretation of stateful training as sampling from an induced memory-state distribution, we do not establish formal convergence guarantees or characterize the stability properties of this distribution. A more rigorous theoretical analysis of the recurrent dynamics remains an important direction for future work. Second, our empirical evaluation is confined to embodied navigation tasks. Although the proposed stateful optimization protocol is general and applicable to sequence models with explicit recurrent states, its effectiveness in other domains such as language modeling or long-horizon control remains to be systematically validated.

% While StateLinFormer demonstrates consistent advantages under persistent memory training, several limitations remain. First, our evaluation is restricted to embodied navigation, and the generality of stateful optimization beyond architectures with explicit recurrent memory remains to be validated. Second, we do not provide a formal theoretical characterization of the induced memory-state distribution or its long-term stability, leaving deeper analytical understanding for future work.

%\section{CONCLUSIONS}
\section{Conclusion}

We introduced StateLinFormer, a linear-attention navigation model trained under a stateful optimization protocol that preserves memory states across consecutive training batches. While conventional training treats recurrent memory as an inference-time artifact, our approach exposes model parameters to the distribution of accumulated memory states induced by long interaction streams. Empirically, this alignment between training and deployment consistently improves long-horizon adaptation and in-context performance under controlled capacity settings. 

Additionally, our findings suggest that the proposed training paradigm enhances the model's in-context learning (ICL) capabilities. We hope this work encourages further investigation into stateful training dynamics and their broader implications for continual and sequential decision-making systems.

%\section*{ACKNOWLEDGMENT}

%%%%%%%%%%%%%%%%%%%%%%%%%%%%%%%%%%%%%%%%%%%%%%%%%%%%%%%%%%%%%%%%%%%%%%%%%%%%%%%%

\balance
\bibliography{example_paper}
\bibliographystyle{plain}

%%%%%%%%%%%%%%%%%%%%%%%%%%%%%%%%%%%%%%%%%%%%%%%%%%%%%%%%%%%%%%%%%%%%%%%%%%%%%%%%
\newpage
\section*{APPENDIX}
\subsection{Details of Model Structures}

We utilize RWKV-7 \cite{peng2025rwkv} as our linear attention architecture to model state transitions. Specifically, the model is configured with 18 layers and a hidden dimension of 512, totaling approximately 0.2 billion parameters, with 32 attention heads per layer dedicated to processing spatial representations.
For feature extraction, both the goal encoder $\mathrm{E}_{\text{goal}}$ and the image encoder $\mathrm{E}_{\text{image}}$ utilize SigLIP pre-trained weights, while the visual encoder $\mathrm{E}_{\text{visual}}$ follows the architecture design of SPOC. 

\subsection{Details of Training} 
During training, the parameters of $\mathrm{E}_{\text{goal}}$ and $\mathrm{E}_{\text{image}}$ are kept frozen. The training process is initialized with a learning rate of $\mathrm{1e}^{-4}$, which follows a linear decay schedule throughout the training duration. We employ a batch size of 4 for all experiments. The model is trained on 8 NVIDIA A800 GPUs. We train our model for a total of 10 epochs.

\end{document}